\documentclass[letterpaper, 10 pt, conference]{ieeeconf}  

\IEEEoverridecommandlockouts                              

\usepackage{graphics} 
\usepackage{epsfig} 
\usepackage{mathptmx} 
\usepackage{times} 
\usepackage{amsmath} 
\usepackage{amssymb}  
\usepackage{graphicx}
\usepackage{nicematrix}
\usepackage{tikz}
\usepackage{xcolor}
\usepackage{xspace}
\usepackage{makecell}
\usepackage{booktabs} 
\usepackage{tabularx}
\usepackage{multirow}
\usepackage[table]{xcolor}
\usepackage{caption}
\usepackage{array}
\usepackage{booktabs}
\usepackage{dblfloatfix}
\usepackage[hidelinks]{hyperref}
\usepackage[many]{tcolorbox}
\usepackage{cuted}
\usepackage{pifont}

\newcommand{\method}[1]{\textsc{#1}}

\newcommand{\yes}{\ding{51}}        

\newcolumntype{Y}{>{\centering\arraybackslash}X}

\title{\LARGE \bf
DREAM: Deployment-Time Demonstration Generation via Real-to-Sim for Scalable Policy Adaptation
}

\author{
Makoto Sato$^{1}$,
Tatsuya Matsushima$^{1}$,
Yutaka Matsuo$^{1}$,
Yusuke Iwasawa$^{1}$
\thanks{$^{1}$The University of Tokyo, Tokyo, Japan.}%
}

\begin{document}

\maketitle
\thispagestyle{empty}
\pagestyle{empty}


\begin{strip}
\vspace{-5.0em}
\begin{center}
  \begin{minipage}{\textwidth}
    \centering
    \includegraphics[width=\linewidth]{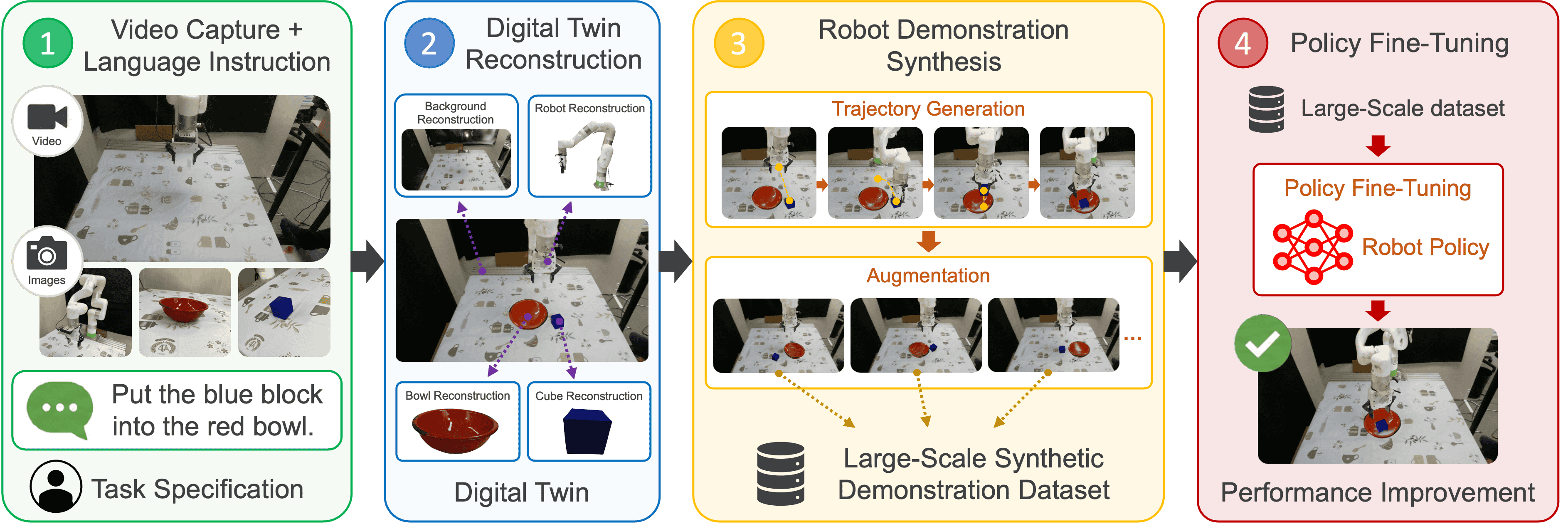}
    \captionof{figure}{\textnormal{\textbf{Overview of \method{DREAM}.} Given a short scene video and a language instruction (e.g., ``put the blue block into the red bowl''), \method{DREAM} reconstructs the workspace as a digital twin, uses task-and-motion planning to automatically generate diverse synthetic image--action data, and fine-tunes a pretrained robot policy that is then deployed in the real workspace. This adapts the policy to the target workspace without requiring any teleoperated demonstrations.}}
    \label{fig:teaser}
    \vspace{-1em}
  \end{minipage}
\end{center}
\end{strip}


\begin{abstract}
Vision--language--action (VLA) models have made strong progress in language-conditioned robot manipulation, but improving their performance in a new workspace still often requires action-labeled data from that environment.
Collecting such data by human teleoperation is costly, especially when each workspace, object arrangement, or task may require new demonstrations.
We present \method{DREAM}, a framework that generates fine-tuning data for a pretrained VLA from a captured workspace and a language instruction, without requiring a task-specific human demonstration.
\method{DREAM} reconstructs the workspace, automatically translates the instruction into symbolic task goals and success criteria using a large language model, and uses task-and-motion planning to generate feasible robot trajectories.
The planned trajectories are augmented across randomized object configurations, verified by the generated success criteria, and rendered into image--action examples for VLA fine-tuning.
Through real-robot experiments on language-conditioned manipulation tasks, we study whether \method{DREAM} can serve as a scalable data-collection system for the deployment workspace by examining whether fine-tuning on its automatically generated data improves success over direct deployment and how its data-collection cost compares with human teleoperation when adapting a VLA to a new workspace.
\end{abstract}


\section{Introduction}
\label{sec:intro}

Vision--language--action (VLA) models have made substantial progress in language-conditioned robot manipulation. Trained on large-scale robot datasets and vision--language priors, recent models can map visual observations and natural language instructions to robot actions across a range of manipulation tasks \cite{groot,pi05}. Ideally, such pretrained policies would be deployed in new workspaces without additional data collection. In practice, however, out-of-the-box performance can remain insufficient when the workspace differs from the conditions covered by pretraining \cite{molmoact2}. Object instances, object poses, clutter, support surfaces, and robot reachability all affect which actions are executable. When performance is insufficient, improving the policy requires action-labeled supervision that reflects the workspace where the robot will operate.

Obtaining such supervision is costly. Human teleoperation yields high-quality image--action demonstrations, but each new workspace, object arrangement, or task demands additional operator time, physical access to the robot, repeated scene resets, and safety monitoring. Autonomous real-robot data collection improves throughput, yet it still ties up hardware and relies on repeated interaction with the deployment environment. As robots are deployed more broadly, expecting users or developers to collect task-specific demonstrations for every new workspace becomes unrealistic. What is needed is a scalable way to generate action-labeled training data for the target workspace without collecting it manually.

Recent robot data-generation methods reduce the need for large-scale teleoperation. Some methods synthesize imitation learning datasets from a small number of demonstrations, while others generate robot training data from real-world captures or use reconstructed environments for policy improvement and evaluation \cite{mandlekar2023mimicgen,yu2025real2render2real,torne2024rialto,jiang2025gsworld}. These directions reduce manual data collection, but they often still require task-specific seed demonstrations, real-robot interaction, or reward design. They do not fully address our target setting, where a user captures a workspace, specifies a task in language, and obtains fine-tuning data without demonstrating the task.

We introduce \method{DREAM} (\textbf{D}eployment-Time Demonstration Generation via \textbf{Rea}l-to-Si\textbf{m} for Scalable Policy Adaptation), a data-generation framework for improving VLA performance in a deployment workspace. Given a short video of the workspace and a language instruction, \method{DREAM} reconstructs the environment, uses a large language model to automatically derive symbolic task goals and success criteria, and uses task-and-motion planning (TAMP) to generate feasible robot trajectories. The planned trajectories are automatically augmented across diverse object configurations, verified with the generated success criteria, and rendered into image--action examples for fine-tuning a pretrained VLA.

The key idea is to use the reconstructed workspace as a source of executable supervision. The resulting digital twin provides environment-specific context, including the workspace layout and appearance, the deployment camera viewpoints, the support surfaces, and the set of manipulable objects with their poses. TAMP uses this context to produce feasible behavior, including grasp choices, motion segments, placement targets, and subgoal orderings. In this way, \method{DREAM} generates supervision that is grounded in the deployment workspace without requiring a task-specific human demonstration.

We evaluate \method{DREAM} on language-conditioned manipulation tasks in a real robot workspace, in two experiments that together test it as a scalable data-collection system for deployment environments. The first asks whether the reconstructed simulation is valid as a data source, adopting the paired sim-and-real evaluation protocol of SIMPLER~\cite{simpler} to measure how closely policy performance in simulation tracks performance on the real robot. The second measures how real-world policy performance scales with the amount of automatically generated data, and how that scaling compares with human teleoperation as a function of data-collection effort.

In summary, our contributions are:
\begin{itemize}
    \item We propose a scalable data-collection framework that turns a short video of the deployment workspace and a language instruction into robot training data, without any task-specific demonstration.
    \item We validate the reconstructed workspace as a data source by measuring the correlation between simulated and real-robot policy performance.
    \item We show that real-world policy performance improves as more data is generated, and that this improvement requires no additional human
    demonstration.
\end{itemize}
\section{Related Work}
\label{sec:related}

\textbf{Planning-based data generation.}
Planning provides an alternative to direct teleoperation for generating robot demonstrations.
Scaling Up and Distilling Down combines language-based task decomposition with motion and grasp planning to generate demonstrations for language-conditioned policy learning~\cite{ha2023scalingup}.
Sampling-based contact planning has also been used to generate demonstrations for contact-rich manipulation, where trajectory consistency is important for effective policy learning~\cite{zhu2024contactrich}.

For long-horizon manipulation, task-and-motion planning (TAMP) jointly reasons over discrete task sequences and continuous, geometrically feasible robot motions~\cite{garrett2021tamp}.
OPTIMUS uses TAMP-generated demonstrations to train visuomotor policies~\cite{dalal2023optimus}, while HITL-TAMP combines planning with human teleoperation for difficult contact-rich segments~\cite{mandlekar2023hitltamp}.
NOD-TAMP further improves generalization by incorporating neural object descriptors into long-horizon planning~\cite{cheng2024nodtamp}.
However, these methods do not explicitly address the visual sim-to-real gap.
In contrast, \method{DREAM} reconstructs the real environment and generates image--action data that reflect its appearance and camera configuration.

\textbf{Data augmentation from demonstrations.}
Several systems scale imitation learning by expanding one or a few human demonstrations into larger datasets.
MimicGen adapts object-centric motion segments from a small number of source demonstrations to new scene configurations, object instances, and robot embodiments \cite{mandlekar2023mimicgen}.
SkillMimicGen segments demonstrations into reusable skills and stitches them through transit and transfer motions for long-horizon manipulation \cite{garrett2024skillmimicgen}.
DexMimicGen extends this paradigm to bimanual dexterous manipulation, automatically generating large-scale demonstrations across diverse task variations from a small set of demonstrations \cite{jiang2025dexmimicgen}.
DemoGen generates synthetic demonstrations to improve the spatial generalization of visuomotor policies from limited human data \cite{xue2025demogen}.
RoboCasa scales such generation to large simulated datasets of everyday tasks for training generalist robots \cite{robocasa}.
More recently, SoftMimicGen further extends demonstration generation to deformable object manipulation by synthesizing demonstrations across diverse deformable objects, manipulation behaviors, and robot embodiments \cite{moghani2026softmimicgen}.
These approaches substantially reduce the amount of manual data collection, but their generated motions remain conditioned on seed demonstrations.

\textbf{Data generation in reconstructed scenes.} Recent work has used real-world reconstruction to reduce the visual gap between generated data and deployment.
RialTo constructs digital twins from real-world scans and improves imitation policies with reinforcement learning in the reconstructed simulator \cite{torne2024rialto}.
R2R2R reconstructs objects from scans, tracks object motion from a single human demonstration video, and renders robot-agnostic image--action data through inverse kinematics without object dynamics simulation \cite{yu2025real2render2real}.
RoboSplat reconstructs a demonstrated scene with 3D Gaussian Splatting and edits the resulting representation to generate visually diverse demonstrations \cite{robosplat}.
GSWorld studies photorealistic closed-loop simulation by coupling Gaussian-based scene representations with physics engines \cite{jiang2025gsworld}.
RoboSimGS reconstructs real scenes with 3D Gaussian Splatting and generates simulated manipulation data for zero-shot real-world learning \cite{robosimgs}, and a concurrent 3D Gaussian Splatting digital twin renders photorealistic scenes for robotic manipulation \cite{hifidt}.
SplatSim instead uses Gaussian Splatting as a rendering primitive to transfer RGB manipulation policies from simulation to the real world \cite{splatsim}.
In contrast, \method{DREAM} couples real-world scene reconstruction with TAMP-based demonstration generation.
Given a language-specified task, it generates long-horizon image--action demonstrations without requiring a task-specific seed demonstration.

Table~\ref{tab:related_compare} summarizes representative data-generation methods. Existing planning-based generators operate in manually authored simulation environments, while existing real-to-sim pipelines produce behavior by replaying or retargeting seed demonstrations or by task-specific scripted primitives. \method{DREAM} instead plans multi-step demonstrations directly from language-specified goals in a reconstructed deployment workspace, without task-specific seed demonstrations.

\begin{table}[t]
\centering
\caption{
Comparison of representative robot data-generation methods.
\yes{} denotes support, while -- denotes that the capability is not provided.
}
\label{tab:related_compare}

\renewcommand{\arraystretch}{1.10}
\setlength{\tabcolsep}{2.2pt}

\resizebox{\columnwidth}{!}{%
\begin{tabular}{@{}lccccc@{}}
\toprule
Method
& \shortstack{Teleop-\\free}
& \shortstack{Demo-\\free}
& \shortstack{Real-to-\\Sim}
& \shortstack{Language\\task}
& \shortstack{Long-\\horizon} \\
\midrule

Scaling Up~\cite{ha2023scalingup}
& \yes & \yes & -- & \yes & \yes \\

OPTIMUS~\cite{dalal2023optimus}
& \yes & \yes & -- & -- & \yes \\

SkillMimicGen~\cite{garrett2024skillmimicgen}
& -- & -- & -- & -- & \yes \\

RialTo~\cite{torne2024rialto}
& -- & -- & \yes & -- & -- \\

RoboSplat~\cite{robosplat}
& -- & -- & \yes & -- & -- \\

R2R2R~\cite{yu2025real2render2real}
& \yes & -- & \yes & -- & -- \\

GSWorld~\cite{jiang2025gsworld}
& \yes & \yes & \yes & -- & \yes \\

\midrule
\textbf{\method{DREAM}}
& \yes & \yes & \yes & \yes & \yes \\
\bottomrule
\end{tabular}
}

\par\vspace{0.4em}  

\begin{minipage}{\columnwidth}
\scriptsize
\raggedright
\emph{Teleop-free}: robot teleoperation is not required for data generation.
\emph{Demo-free}: no task-specific human or robot seed demonstration is required.
\emph{Real-to-Sim}: real-world capture is used to construct the simulated deployment environment.
\emph{Language task}: language directly specifies the behavior for which demonstrations are generated.
\emph{Long-horizon}: the method supports generating multi-step task demonstrations.
\end{minipage}
\end{table}
\section{Proposed System}
\label{sec:system}

\begin{figure*}[t]
\centering
\includegraphics[width=0.9\textwidth]{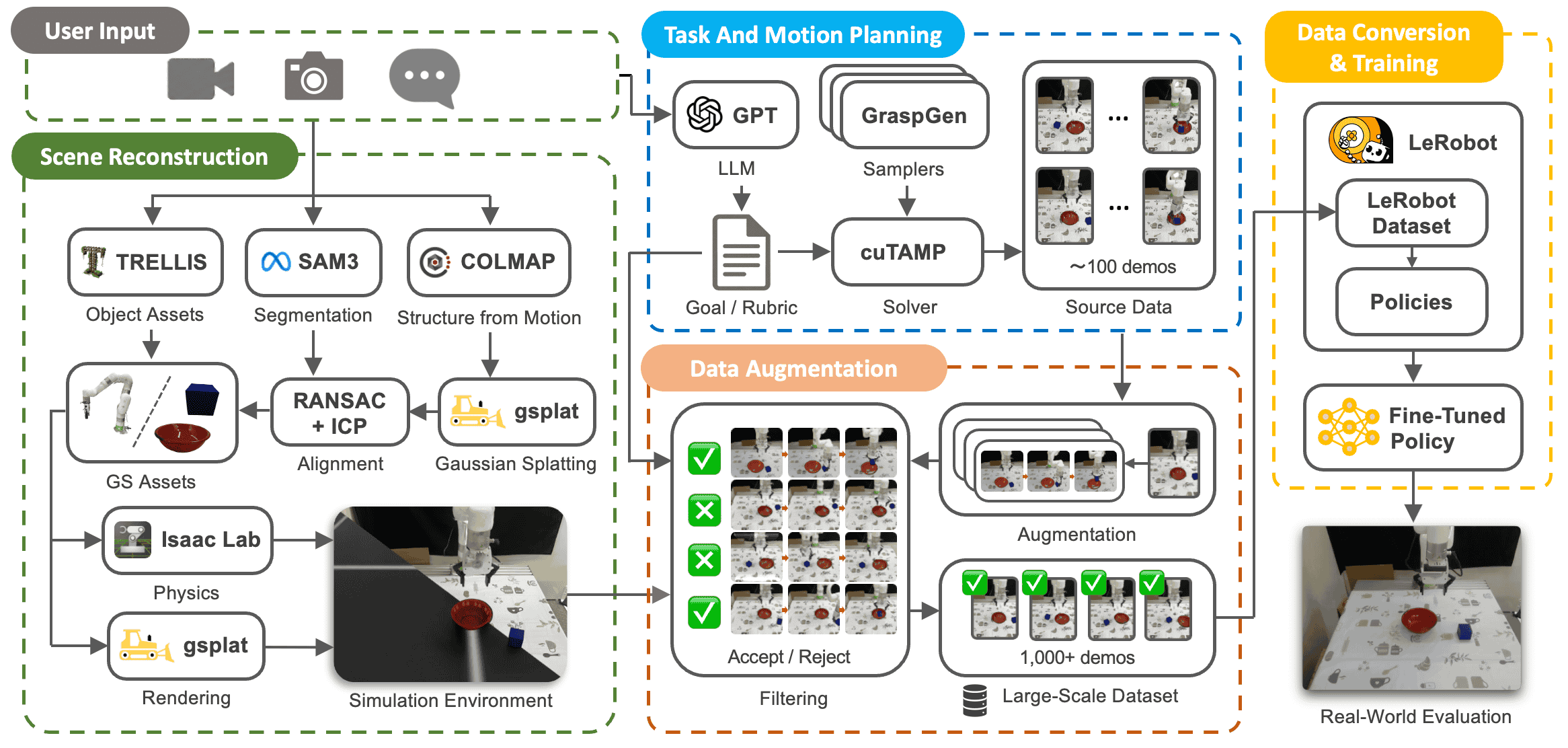}
\caption{
\textbf{The \method{DREAM} pipeline.} Given a short RGB video of the
target workspace and a language instruction, \method{DREAM} reconstructs a
metrically aligned digital twin, translates the instruction into symbolic
goals and staged success criteria with a large language model, generates
feasible robot trajectories with TAMP, augments them across randomized
object configurations, and renders them into photorealistic image--action
demonstrations. The resulting synthetic dataset is used to fine-tune a
pretrained VLA policy, which is then deployed directly in the physical
workspace.
}
\label{fig:pipeline}
\end{figure*}

\method{DREAM} generates deployment-workspace training data without task-specific robot demonstrations, from a short video of the target workspace, a language instruction, and the robot model with its camera calibration.
We assume a static workspace, rigid objects, a recorded capture-time joint configuration, and known extrinsics for the two deployment cameras---one fixed and one wrist-mounted---so that rendered observations match the cameras the policy sees at test time.
The pipeline then proceeds in four stages (Fig.~\ref{fig:pipeline}): digital-twin reconstruction, demonstration synthesis with task-and-motion planning (TAMP), demonstration augmentation, and rendering and dataset construction, followed by policy fine-tuning.

The central design principle is to separate visual realism from physical feasibility.
The reconstruction is used only for rendering, while physics and planning operate on a compact geometric world containing only the robot, the movable objects, and a support surface.
This yields image--action data that is both visually grounded in the deployment workspace and geometrically feasible for the target robot, without requiring watertight scene meshes.

\subsection{Digital-Twin Reconstruction}

\textbf{Capture and splat training.}
Given a short RGB video of the workspace captured by a commodity hand-held camera, \method{DREAM} uniformly extracts frames and estimates camera poses and a sparse point cloud by structure from motion with COLMAP~\cite{colmap}.
A 3D Gaussian Splatting~\cite{gaussiansplatting} representation of the scene is then optimized with gsplat~\cite{gsplat}, honoring the full calibrated intrinsics and disabling world-space normalization so that the reconstruction can subsequently be expressed metrically in the robot frame.

\textbf{Alignment to the robot base frame.}
Monocular reconstruction yields an arbitrary, non-metric coordinate frame, so the scene must be registered to the robot before planning.
Rather than relying on fiducial markers or manually specified correspondences, \method{DREAM} recovers the similarity transform $\mathrm{SIM}(3)$ automatically from the robot's own appearance in the scan.
A reference point cloud is synthesized by posing the robot model with forward kinematics at the recorded capture-time joint configuration.
Robot points in the reconstruction are identified by text-prompted segmentation (SAM3~\cite{sam3}) with multi-view voting across each point's feature track, which suppresses single-view mask leakage.
The two clouds are coarsely registered with FPFH features, RANSAC, and a scale-aware estimator, then refined by a scaled ICP that re-solves a least-squares similarity fit~\cite{umeyama} at each iteration.
The estimated transform is baked into the splat and serialized, so that every later stage operates in the metric robot frame.

\textbf{Asset decomposition.}
A single scan captures one static configuration, whereas data generation requires a moving robot and re-placed objects.
The aligned splat is therefore factorized into independently posable assets.
Gaussians near the posed robot surface are removed, yielding a robot-free static background (the movable objects are absent during capture).
Per-link robot splats are extracted by assigning Gaussians to their nearest link surface and transforming them into link-local frames, producing a splat rig that can be re-posed by forward kinematics.
Each manipulable object is represented by an appearance splat generated from a few photographs by an image-to-3D generative model (TRELLIS~\cite{trellis}) and calibrated to the physical object's metric size, paired with a collision mesh for physics and planning.

\textbf{Environment instantiation.}
The assets are assembled with a lightweight scene-composition tool into a scene description that records the movable-object set, their initial poses, a randomization region, the language instruction, and the camera calibration.
This metadata alone instantiates an IsaacLab~\cite{isaaclab} environment, with no per-task code.
Only a ground plane, the robot articulation, and the movable objects are simulated; the reconstructed surroundings appear exclusively as the Gaussian-splat background.
The fixed and wrist cameras are instantiated from the same calibration used at deployment, so simulation and reality share one camera model.

\subsection{Demonstration Synthesis Using TAMP}

\textbf{Task specification.}
Planning needs a symbolic goal and recording needs a success test, and both are written in a vocabulary we fix in advance rather than one the language model invents.
Each predicate is declared with its name, arity, and argument types, together with an evaluation function over object poses. \texttt{On(x, y)}, for instance, holds when $x$ is aligned with $y$ in the horizontal plane and sits at the height implied by the two objects' extents.
The role of the model (GPT-4o-mini) is therefore compositional. It selects and instantiates predicates whose semantics are already implemented, and never supplies semantics of its own.

From the instruction, the model emits goal atoms over that vocabulary and the objects present in the scene (e.g., ``put the fruits on the plate'' $\rightarrow$ \texttt{On(banana, plate)}, \texttt{On(peach, plate)}).
Each response is parsed by the grammar of the predicate it names and checked for arity and for arguments that resolve to real scene objects, so malformed output fails closed rather than corrupting the task.
Success criteria are emitted in the same way, as a \emph{staged rubric}: an ordered sequence of stages (e.g., reach, grasp, transport, place), each naming one of thirteen predefined checkers with typed arguments and optional failure conditions, compiled and validated against the scene before use.
Because the model can only select and parameterize implemented predicates, specification errors manifest as missing demonstrations rather than as incorrect action labels.

The initial symbolic state is derived from the scene rather than from the model. Each movable object is assigned the support it rests on by comparing object poses and extents, and the remaining atoms---that the hand is empty, and which objects are movable or surfaces---follow from the scene description.

\textbf{Planning.}
Planning proceeds in two stages.
Plan skeletons---operator sequences over a PDDL domain (\texttt{MoveFree}, \texttt{MoveHolding}, \texttt{Pick}, \texttt{Place}, \texttt{Push}) whose continuous parameters remain free---are enumerated by breadth-first search over grounded operators.
Following cuTAMP~\cite{cutamp}, the free parameters of each skeleton (grasps, placements, robot configurations) are resolved by GPU-parallel optimization over a batch of 1000 particles: particles are initialized from sampling streams---a diffusion-based grasp generator (GraspGen~\cite{graspgen}), and a heuristic placement sampler over the support surface---and descended on a weighted constraint cost covering collision, reachability, and kinematic tolerances.
The best satisfying particle is refined into an executable trajectory by segment-wise, collision-aware joint-space motion planning with cuRobo~\cite{curobo}.

\textbf{Execution and recording.}
Each plan is executed open-loop in the simulator, monitored by a detector that aborts the episode when the gripper closes without acquiring the object.
Episodes that pass the rubric are recorded as time-indexed trajectories of robot states, gripper commands, and object states, with every frame annotated by the symbolic operator it belongs to---an alignment available precisely because the demonstrations come from a planner.
The robot starts every episode from the same fixed home configuration, while object placements and planner samples are randomized per episode to diversify the source pool, and unlike replay-based generators, no seed demonstration of the target task is required.

\subsection{Demonstration Augmentation}

TAMP demonstrations are physically valid but expensive, each requiring a planning call and a rollout.
\method{DREAM} therefore multiplies a small source pool into a large dataset with MimicGen-style augmentation~\cite{mandlekar2023mimicgen}.
Because the source trajectories carry their symbolic plan, subtask boundaries are read directly from the plan rather than inferred heuristically from the signals; boundaries are shifted past the post-grasp lift, and terminal settle and homing motions are trimmed.
For each new trial, object poses are re-randomized, a source demonstration is selected per subtask by nearest neighbor in object pose, its end-effector segment is transformed rigidly into the new object frame, and consecutive segments are connected by interpolated bridge motions.
A trial is accepted only if the rubric's terminal criterion fires, and accepted demonstrations additionally pass an offline physical-plausibility filter that rejects disturbance of static objects, pre-grasp bumping of the target, dragging at support height, and joint overspeed.

\subsection{Rendering and Dataset Construction}

Augmentation produces no images; observations are attached afterwards by replaying the recorded states through the Gaussian-splat scene.
For each frame, the robot-free background, the per-link robot splats posed by forward kinematics from the recorded joint trajectory, and the object splats posed by the recorded object poses are concatenated into a single Gaussian set, and RGB observations are rasterized for the policy's cameras: the fixed third-person camera at its calibrated extrinsics, and the wrist camera posed by forward kinematics through its hand-eye calibration.
Because all assets live in one Gaussian set, mutual occlusion between the robot, the objects, and the scene is resolved implicitly by depth-sorted alpha blending; no masks, depth comparisons, or ray-traced passes are required.

The final dataset stores, at each timestep, the language instruction, the rendered $640\times480$ RGB observations, robot states, and action labels, and is written in the LeRobot~\cite{cadene2026lerobot} dataset format, from which the input--output representation required by each target policy is derived.

\subsection{Policy Fine-tuning}

Finally, \method{DREAM} uses the synthesized demonstrations to train robot policies via imitation learning. The trained policies are deployed in the corresponding real-world environment, where they receive real camera observations and robot states in the same format used during training. Language-conditioned policies, such as VLA models, additionally receive the task instruction. TAMP is used only during data generation; at deployment, the policy predicts actions directly from its observations, without access to the planner.

\section{Experiments}
\label{sec:experiments}

Our experiments evaluate whether \method{DREAM} can serve as a scalable data-collection system for a previously unseen deployment workspace. We organize the evaluation around two research questions.

\noindent\textbf{RQ1 (reliability of the reconstructed environment):}
Can the digital twin reconstructed by \method{DREAM} serve as a trustworthy environment for data generation and policy evaluation, i.e., does policy performance measured in the reconstructed simulation correlate with performance in the corresponding real-world workspace?

\noindent\textbf{RQ2 (scalability as a data-collection system):}
Can \method{DREAM} substitute for manual data collection in the deployment workspace, i.e., do its automatically generated demonstrations improve real-world policy performance, and how efficiently does performance scale with human effort compared with teleoperation-based data collection?

\subsection{Experimental Setup}
\label{sec:exp_setup}

\begin{figure}[t]
    \centering
    \begin{tabular}{@{}c@{\hspace{0.02\columnwidth}}c@{}}
        \text{Initial State (Sim / Real)} & \text{Target State} \\[3pt]
        \includegraphics[width=0.47\columnwidth]{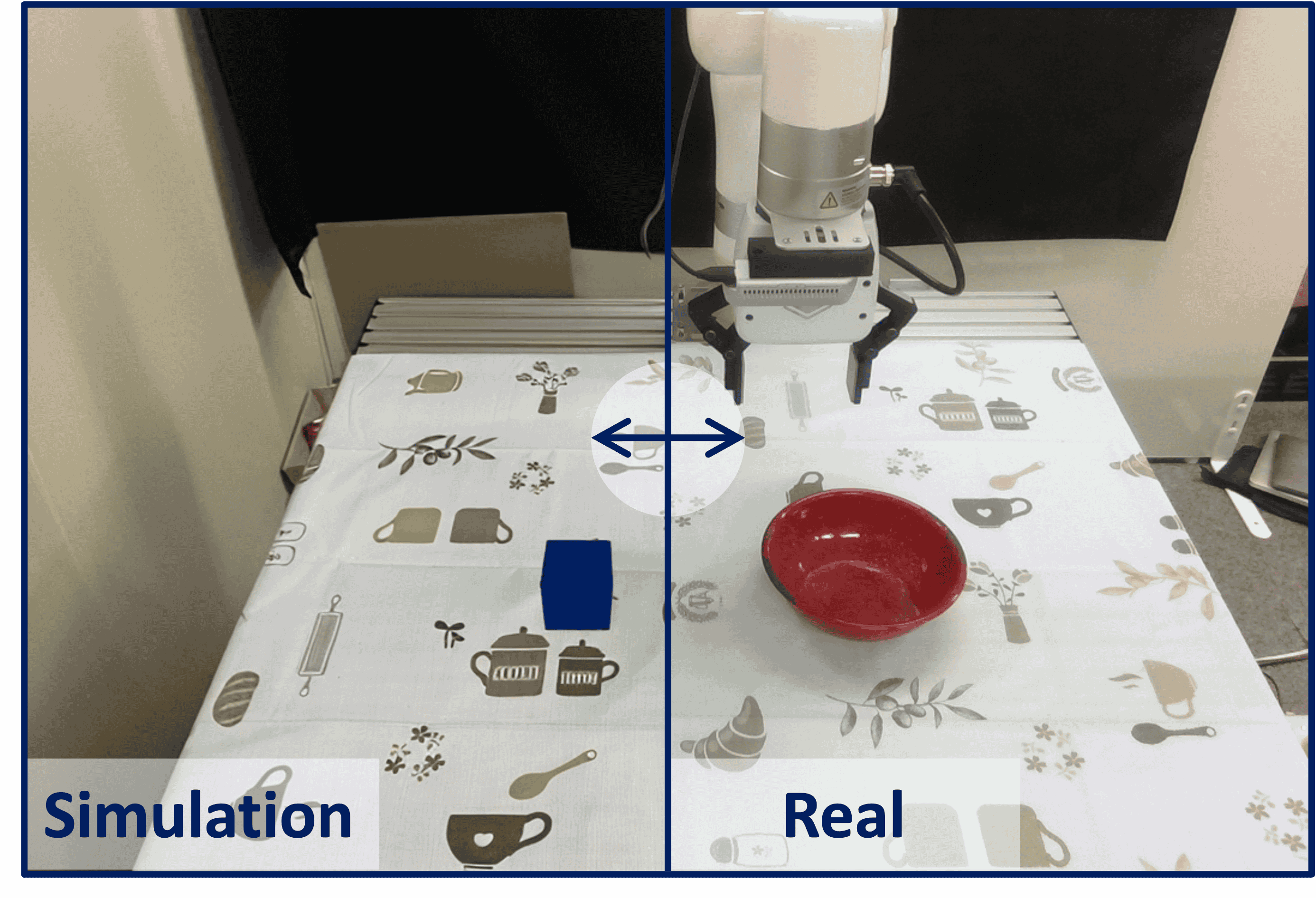}
        &
        \includegraphics[width=0.47\columnwidth]{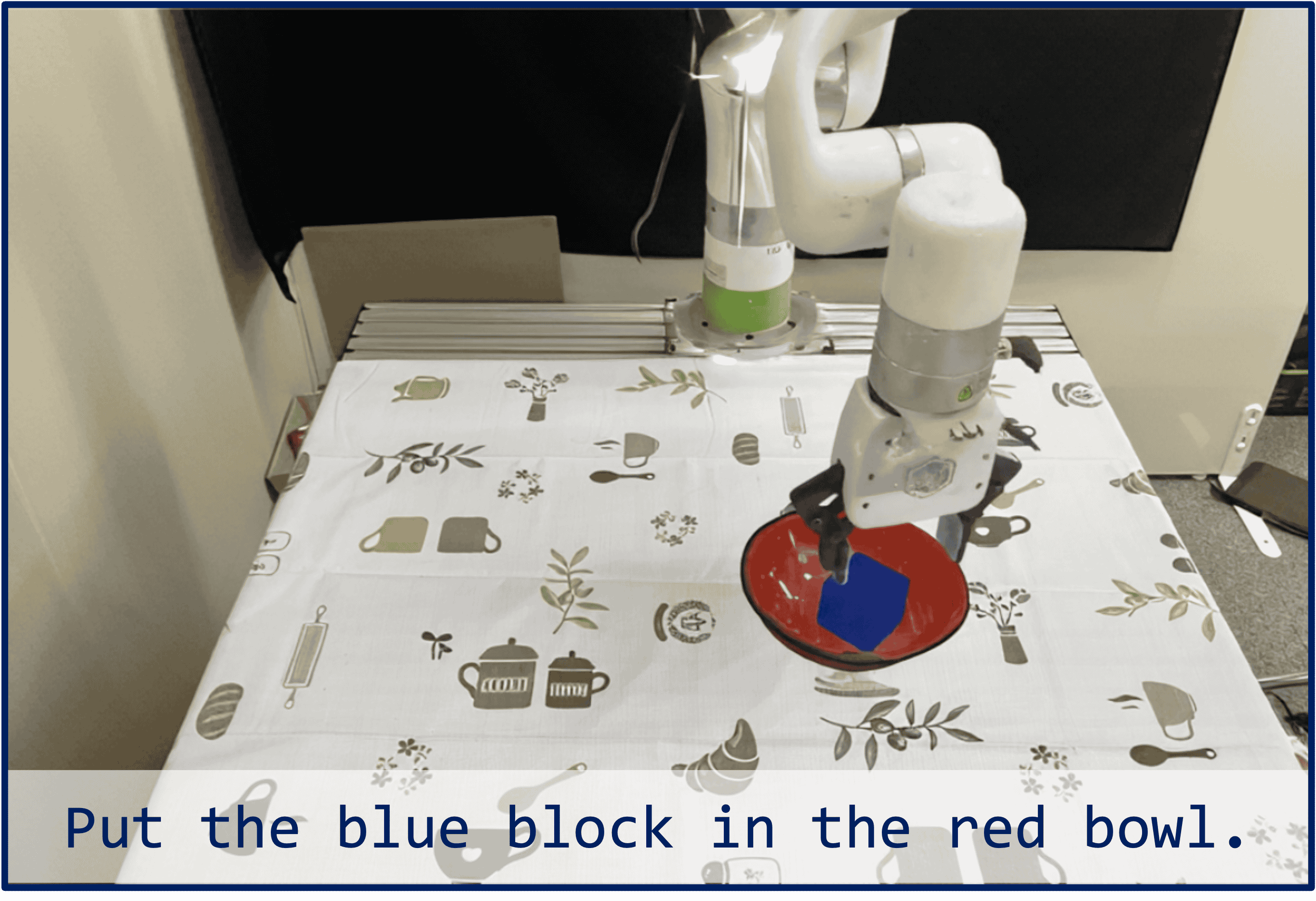}
        \\[-1pt]
        \multicolumn{2}{c}{\text{(a) \emph{BlockIntoBowl}}}
        \\[6pt]
        \includegraphics[width=0.47\columnwidth]{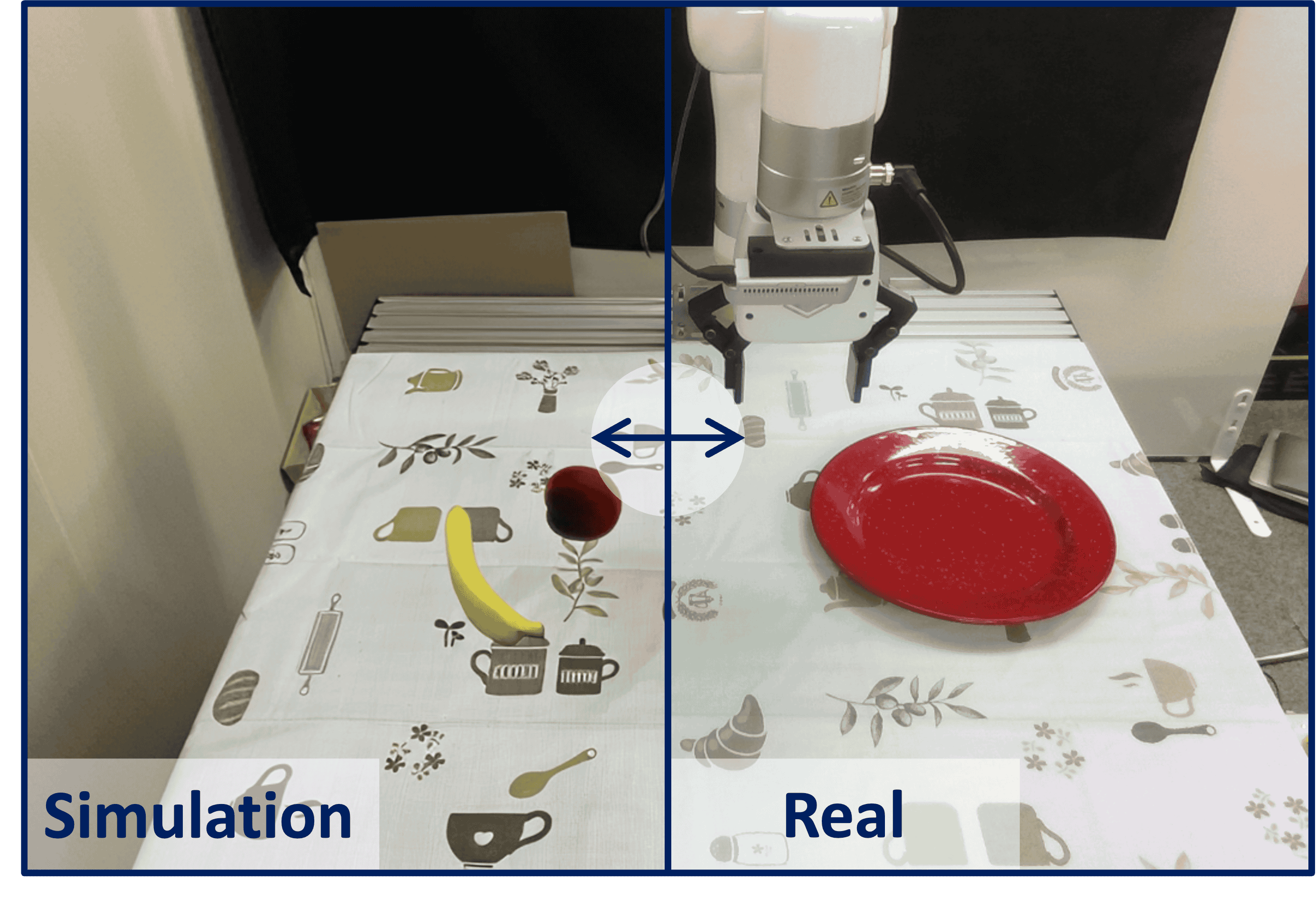}
        &
        \includegraphics[width=0.47\columnwidth]{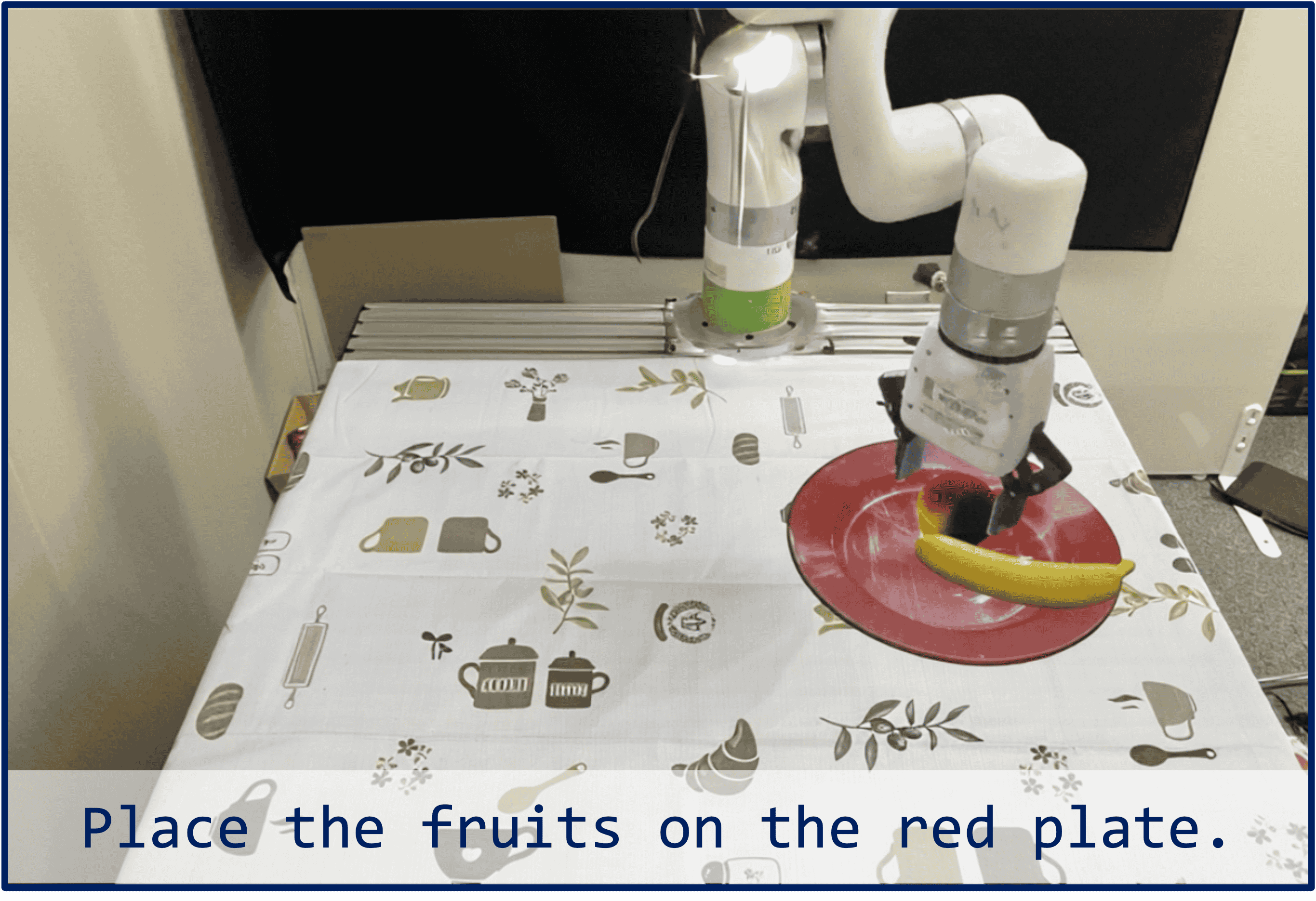}
        \\[-1pt]
        \multicolumn{2}{c}{\text{(b) \emph{FruitPacking}}}
    \end{tabular}
    \caption{
        The two tabletop manipulation tasks. For each task, the left image compares the initial state in the reconstructed digital twin (Simulation) with the corresponding real workspace (Real), and the right image shows the target state, rendered in the reconstructed simulation, together with the language instruction. In \text{(a) \emph{BlockIntoBowl}}, the robot places a blue block into a red bowl. In \text{(b) \emph{FruitPacking}}, the robot places a banana and a peach onto a red plate in sequence.
    }
    \label{fig:tasks}
    \vspace{-3mm}
\end{figure}

\textbf{Tasks.}
We evaluate on two tabletop manipulation tasks using an xArm7 robot, as shown in Fig.~\ref{fig:tasks}.
\emph{BlockIntoBowl} is a single-step pick-and-place task in which the robot places a blue block into a red bowl.
\emph{FruitPacking} requires the robot to place a banana and a peach onto a red plate in sequence; this multi-step task tests whether \method{DREAM} extends to longer-horizon manipulation.

\textbf{Hardware.}
Real-world observations are captured by two Intel RealSense cameras: a D435i mounted at a fixed external viewpoint overlooking the workspace and a D435i mounted on the robot wrist.
The extrinsic parameters of both cameras---the external camera with respect to the robot base and the wrist camera with respect to the end-effector---are calibrated in advance using ArUco markers, and the same camera configuration is reproduced in the reconstructed simulation when rendering training observations.
Both camera views are provided to the policies as visual observations.

\textbf{Policies.}
We evaluate two VLA policies, SmolVLA~\cite{smolvla} and $\pi_{0.5}$~\cite{pi05}, both fine-tuned from their pretrained initializations using supervised imitation learning.
For the sim-to-real correlation analysis in Sec.~\ref{sec:exp_sim_real}, we additionally train Diffusion Policy~\cite{diffusionpolicy} from scratch, to obtain policy checkpoints covering a wider range of performance.
All policies receive RGB observations and the proprioceptive states required by their respective architectures, and the VLA policies additionally receive a language instruction.

\textbf{Data Collection.}
We compare policies trained on demonstrations obtained from two sources:
\begin{itemize}
\item \textbf{Human teleoperation}: demonstrations collected directly in the real environment using a kinematics-based leader--follower teleoperation system. A human operator controls a LeRobot SO-101 leader arm~\cite{cadene2026lerobot}, whose end-effector motion is retargeted to an xArm7 follower through forward and inverse kinematics. RGB observations, proprioceptive states, and robot actions are recorded during each rollout.
\item \textbf{\method{DREAM}}: demonstrations generated automatically in simulation. TAMP-generated source trajectories are synthesized in a digital twin reconstructed from a capture of the deployment workspace, augmented using MimicGen-style augmentation, and converted into image--action demonstrations through Gaussian-splat rendering.
\end{itemize}
Human teleoperation provides demonstrations collected through physical robot execution, whereas \method{DREAM} generates demonstrations entirely in simulation without executing each training trajectory on the physical robot. Policies trained on both data sources are evaluated on the physical robot under the same test conditions.

\textbf{Metrics.}
Our primary metric is task success rate, measured over 15 trials per policy and task under held-out object configurations, where a trial is successful when the final scene satisfies the task-success criterion.
To characterize the generator, we report valid trajectory ratio, planning success rate, augmentation acceptance rate, and average planning time per valid trajectory.
To compare cost, we measure human-effort time---workspace capture, calibration and alignment, and task specification for \method{DREAM}, versus teleoperation time for human data---and report \method{DREAM}'s automatic generation time separately, since the planning, augmentation, and rendering stages run without human attention once the task specification and success criteria are defined.

Policy training was performed using NVIDIA GB200 GPUs. All other computational stages, including Gaussian Splatting optimization, object and robot asset generation, demonstration generation, simulation, and rendering, were executed on three NVIDIA RTX A6000 GPUs with 48 GB of memory each.

\subsection{Experiment 1: Sim-to-Real Correlation}
\label{sec:exp_sim_real}

\begin{figure}[t]
    \centering
    \includegraphics[
        width=\columnwidth,
        trim=0 0 0 0,
        clip
    ]{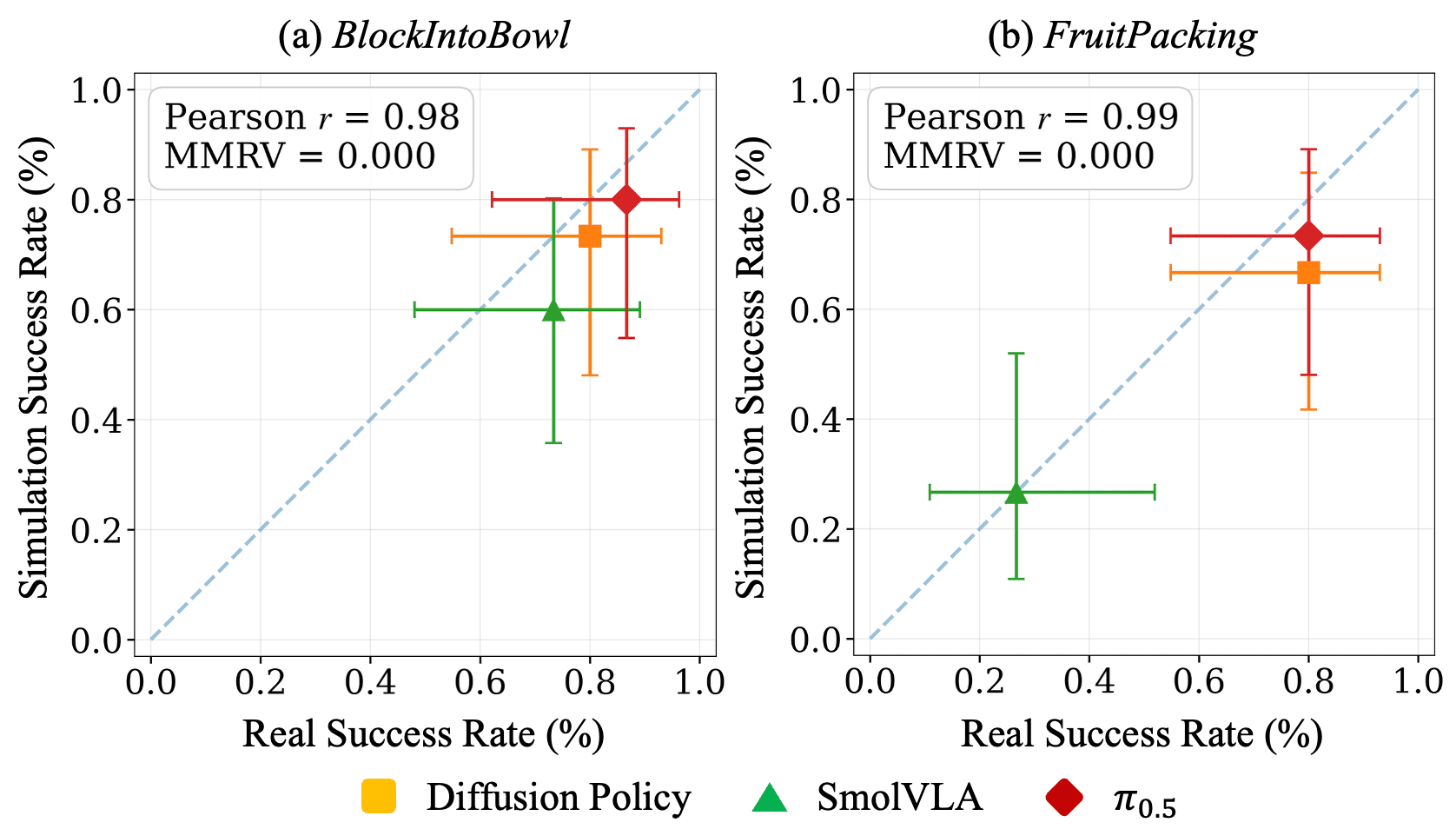}
    \caption{
        Sim-to-real correlation of policy performance.
        Each point represents a policy checkpoint trained with a particular dataset size and random seed.
        The horizontal axis shows real-world success rate, and the vertical axis shows success rate in the reconstructed simulation.
        Colors distinguish the two tasks, while marker shapes distinguish the three policy architectures
        (Diffusion Policy, SmolVLA, and $\pi_{0.5}$).
        The dashed line represents equal simulated and real-world performance.
    }
    \label{fig:sim_real_correlation}
    \vspace{-3mm}
\end{figure}

Experiment~1 addresses RQ1 by asking whether policies can be compared inside the reconstruction.
We train checkpoints of each architecture on the 100 teleoperated demonstrations collected for each task and run the same checkpoints in the reconstruction and on the robot under matched object configurations and identical success criteria.
Figure~\ref{fig:sim_real_correlation} plots simulated against real success rate, one panel per task and one marker per policy, with the dashed line marking equal performance.
Agreement is summarized by Pearson's $r$ and the mean maximum rank violation (MMRV)~\cite{simpler,polaris}, the latter being zero when the reconstruction never orders a pair of policies differently from the robot.

Both tasks give $\mathrm{MMRV} = 0.000$, with $r = 0.98$ on \emph{BlockIntoBowl} and $r = 0.99$ on \emph{FruitPacking}, so the reconstruction reproduces the real ranking of the three policies in both cases.
Every marker lies on or below the diagonal. Simulated success is 7 to 14 percentage points lower than real success, the sole exception being SmolVLA on \emph{FruitPacking}, where the two coincide at the low end of the range.
The reconstruction is thus pessimistic about absolute performance while preserving the comparison between policies, which is the property Experiment~2 relies on when it uses the reconstruction to produce training data.

\subsection{Experiment 2: Scaling with Data Collection Time}
\label{sec:exp_scaling}

\begin{figure}[t]
    \centering
    \includegraphics[width=0.90\columnwidth]{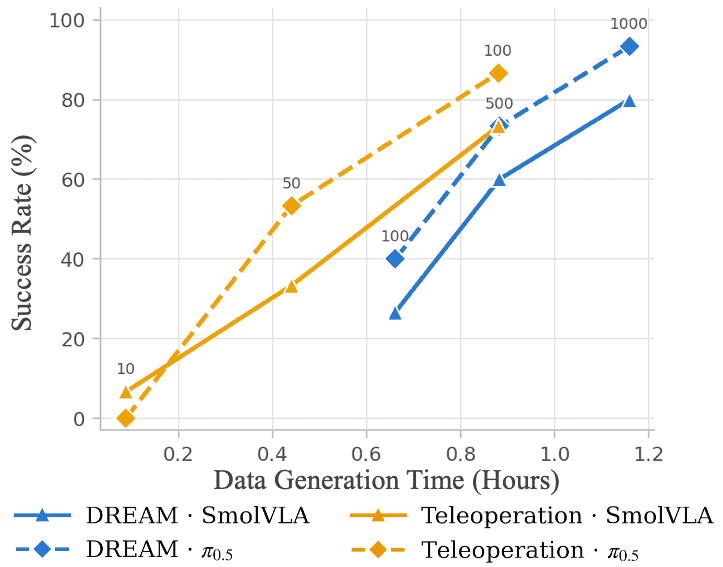}
    \includegraphics[width=0.90\columnwidth]{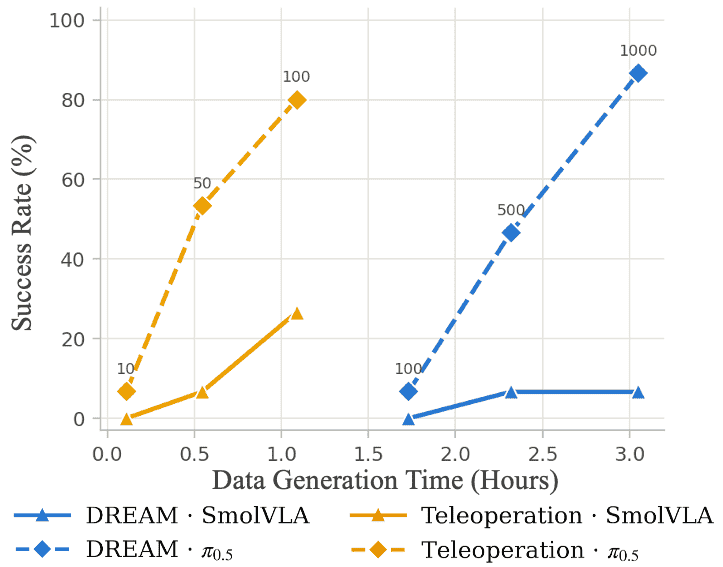}
    \caption{
        Real-world success rate as a function of cumulative data acquisition time for \emph{BlockIntoBowl} (top) and \emph{FruitPacking} (bottom). Each panel compares SmolVLA and $\pi_{0.5}$ trained on demonstrations obtained through human teleoperation or generated by \method{DREAM}. Human-teleoperation datasets contain 10, 50, and 100 demonstrations, whereas \method{DREAM} datasets contain 100, 500, and 1000 demonstrations.
    }
    \label{fig:performance_scaling}
    \vspace{-3mm}
\end{figure}

Experiment~2 addresses RQ2 by asking what a demonstration costs and what it buys.
Teleoperation provides 100 demonstrations per task in nested subsets of $10$, $50$ and $100$; \method{DREAM} provides 1000 in subsets of $100$, $500$ and $1000$.
Each subset fine-tunes SmolVLA and $\pi_{0.5}$, evaluated on the physical robot and plotted against acquisition time in Fig.~\ref{fig:performance_scaling}.
\method{DREAM} is charged one minute of capture, the reconstruction and the 50 TAMP source demonstrations as a fixed cost, then augmentation and rendering per generated demonstration, while teleoperation is charged the operator time each demonstration takes.
Dataset writing is excluded from both.

With enough generated data both tasks end above teleoperation.
On \emph{BlockIntoBowl}, $\pi_{0.5}$ reaches $93.3\%$ from 1000 generated demonstrations against $86.7\%$ from 100 teleoperated ones, and SmolVLA reaches $80.0\%$ against $73.3\%$.
On \emph{FruitPacking}, $\pi_{0.5}$ reaches $86.7\%$ against $80.0\%$, but SmolVLA stalls at $6.7\%$ against $26.7\%$, and adding demonstrations beyond 500 does not help it.

The two routes are priced differently.
Teleoperation stays linear, paying operator time for every demonstration it ever produces, whereas \method{DREAM} pays 36 and 95 minutes before its first demonstration exists and a few seconds for each one after that, so generating becomes cheaper in wall-clock beyond 73 and 168 demonstrations.
Almost all of that fixed cost is TAMP, and it grows with the horizon.
Collecting the 50 source demonstrations takes $25.8$ minutes on \emph{BlockIntoBowl} and $84.6$ on \emph{FruitPacking}, because episodes are $2.7\times$ longer and the planner's success rate falls from $100\%$ to $62\%$.
Augmentation supplies the rest of the cost, and is dominated by rejection rather than by simulation speed, with only $33.0\%$ and $32.6\%$ of trials passing the rubric and plausibility filters.

A generated demonstration is therefore worth less than a teleoperated one and costs an order of magnitude less to make, and what it costs is machine time.
The gap at equal dataset size comes from the data itself, as the augmented trajectories contain frequent wrist rotations and slightly redundant motions introduced by randomization, so the generated data improves the policy only once it is scaled well past the teleoperated count.
This is what makes \method{DREAM} scalable as a data-collection system.
Enlarging the dataset consumes machine time that grows far more slowly than operator time, while the human contribution stays one scan, one calibration and one instruction whether the dataset holds 100 demonstrations or 1000, so the dataset can be pushed until the policy is good enough rather than until the operator stops.

\section{Conclusion}
\label{sec:conclusion}

We presented \method{DREAM}, a deployment-time framework that turns a captured workspace and a language task specification into workspace-specific fine-tuning data for vision--language--action policies.
By reconstructing an interactable digital twin and using task-and-motion planning to synthesize and validate feasible trajectories, \method{DREAM} adapts a pretrained VLA to a target environment without any human demonstration of the task.

Our experiments are designed to test whether \method{DREAM} can serve as a scalable data-collection system for the deployment workspace, evaluating whether planning-generated demonstrations improve workspace-specific policy performance while reducing task-specific human data collection.
The current framework targets approximately static tabletop scenes with rigid objects; extending it to deformable and articulated objects and more contact-rich tasks is an important direction for future work.
We believe that using reconstructed deployment scenes as a source of action-labeled supervision is a promising route to more scalable VLA deployment.


\bibliographystyle{IEEEtran}
\bibliography{ref.bib}

\end{document}